# A virtual reality-based method for examining audiovisual prosody perception

Hartmut Meister[1], Isa Samira Winter[1], Moritz Wächtler[1], Pascale Sandmann[2], Khaled H. A. Abdellatif[1]

[1]Jean Uhrmacher Institute for Clinical ENT Research, University of Cologne

[2]University of Cologne, Faculty of Medicine and University Hospital Cologne, Department of Otorhinolaryngology, Head and Neck Surgery, Audiology and Pediatric Audiology, Cochlear Implant Center

Corresponding author: Hartmut Meister (hartmut.meister@uni-koeln.de)

**Abstract**

Purpose: Prosody plays a vital role in verbal communication. Acoustic cues of prosody, such as intonation patterns, intensity, and duration have been extensively examined. However, prosodic characteristics are not only perceived auditorily, but also visually, as head and facial movements accompany prosody production. The purpose of this technical report is to present a method for examining audiovisual prosody based on virtual reality.

Methods: While controlled manipulations of acoustic cues are a valuable method for uncovering and quantifying prosody perception, such an approach for visual prosody is much more complicated. Here, we describe a novel approach based on video-realistic animations of a virtual human. Such a method has the advantage that – in parallel to acoustic manipulations – head and facial movements can be parametrized.

Results: We show that animations based on a virtual human provide motion cues similar to those obtained from video recordings of a real talker. Parametrization of the motion cues allows fine-grained manipulations of the movements.

Conclusions: The use of virtual reality promises to open up new avenues for examining multimodal effects of verbal communication. Specifically, we discuss the possible application of this technique in the framework of examining prosody perception in listeners with cochlear implants.

Key words: virtual reality, language, prosody, audiovisual, cochlear implants

**Introduction**

Spoken language is based on segmental features such as vowels and consonants. They enable recognizing words and sentences and allow comprehending a message. However, prosodic features – also denoted as suprasegmentals - play an important role as well. They extend over more than one segment and are thus established by a comparison of items in a sequence rather than in an isolated manner (Lehiste, 1970). Two types of prosody are typically distinguished, namely affective and linguistic prosody (cf., Belyk & Brown, 2014). Affective prosody relates to the expression of emotions and thus gives insight into the speaker's emotional state. In contrast, linguistic prosody provides cues regarding syntax and pragmatics. For instance, it gives information whether an utterance is declarative or an interrogation (i.e., statement vs. question). Moreover, it enables stressing single words in a phrase by giving a narrow instead of a broad focus consequently helping to express a specific meaning.

From an acoustic standpoint, three characteristics are of particular importance for prosodic features, namely intonation, intensity, and duration. For instance, stressed elements typically exhibit a higher voice fundamental frequency (F0), a higher intensity, and a longer duration (Nooteboom et al., 1978). Thus, prosody is related to the melodic and rhythmic structure of an utterance, reflecting its etymology, i.e. the Greek word "prosoidia", meaning 'song sung to music'.

These acoustic features of prosody are typically relatively robust. For instance, van Zyl & Hanekom (2011) have shown that sentence recognition was significantly more prone to background noise than prosody perception. However, prosody perception may be much more problematic for listeners with hearing impairment, especially in individuals with severe hearing loss using cochlear implants (CI) that enable understanding speech by electrical stimulation of the auditory nerve (e.g., Chatterjee & Peng, 2008; Meister et al., 2009; Marx et al., 2015).

The importance of different acoustic cues for the perception of prosody is widely explored. Particularly the option of a targeted manipulation of isolated cues and assessment of corresponding changes in perception helped a lot in order to uncover the underlying mechanisms (e.g., Tao et al., 2006; Meister et al., 2011). However, prosody is not only transmitted via the auditory system but also visually (see overview in Krahmer & Swerts, 2009). This is partly due to the simple fact that changes in acoustic features are produced by changes in articulatory movements. As a result, stressed syllables are typically associated with larger mouth openings (Scarborough et al., 2009).

Not only articulatory movements but also head and eyebrow movements are associated with the production of prosody, especially with changes in F0 (e.g., Munhall et al., 2004; Granström & House, 2005). More recently, Cvejic et al. (2012) have shown that such movements can be used to identify prosodic features. Their participants were able to match the visual information with appropriate acoustic prosody cues, even when only the upper part of the talker's face was displayed. In a follow-up study from the same lab, Kim et al. (2014) tracked eyebrows and head movements associated with prosody production by means of optical markers. By temporarily aligning these movements with the time course of the acoustic speech signal, they were able to identify several distinct patterns. Thus, face and head movements give important information for prosody perception, though the movements may be individually different (Wagner et al., 2014, Kim & Davis, 2016).

Taken together, it is evident that the perception of prosody is multimodal. Everyday communication typically gives access to both acoustic and visual features. Hence, in terms of ecological validity it may not be sufficient to examine prosody perception exclusively in a single modality. However, in contrast to *acoustic* features, the targeted manipulation of *visual* features is imposed to methodological constraints, given the complex movements of head and face. Here, advances in virtual reality techniques, which become increasingly popular, might be a solution. Recent examples are the Metahuman Creator™ within the framework of the Unreal Engine™, the Omniverse audio2face™ application by Nvidia™, or Reallusion's iClone™ software. Such applications allow visualizing arbitrary video-realistic facial, head, and even body movements and thus may represent valuable research tools. However, their use in studying verbal communication, particularly prosody, is still in its infancy. One early example is the work by Ding et al. (2013) who animated rigid head and eyebrow movements accompanying affective prosody. More recent work presents the development of video-realistic speech synthesis with lip-syncing and expression capabilities (e.g., Filntisis et al., 2017; Aneja et al., 2019). In particular, the approach by Aneja et al. (2019) is based on a video of a real talker used as input. Here, a face detector provides several landmarks that are dynamically used to synthesize the expressions on a virtual human's face.

The present report describes procedural and technical aspects of such an approach in the framework of investigating multi-modal features of linguistic prosody. As an example we show the animation of head and eyebrow movements during different narrow focus conditions. In accordance with Aneja et al. (2019) the application is based on two stages. The first one considers techniques of motion capture ("mocap") by the use of state-of-the-art face recognition algorithms. The second stage is the application of virtual humans with a realistic appearance. These virtual humans can be controlled by mocap data and – even more importantly – most of the underlying parameters can be modified arbitrarily. In accordance with the modification of acoustic cues this may give the opportunity to manipulate visual prosodic features. Here, we describe the technical evaluation of such an approach and discuss the potential application of the method in the framework of examining audiovisual prosody perception in CI users.

**Methods**

*Capturing natural head and face movements*

In order to capture natural head and face movements, a dialogue situation between two interlocutors was realized. This approach is based on a procedure described by Cvejic et al. (2010). The task of interlocutor 1 is to provoke natural prosodic expressions of interlocutor 2 during a realistic face-to-face communication. Movements of interlocutor 2 are video-recorded and applied as stimuli.

For the purpose of the present report, the phrase "Der Mann faehrt ein gelbes Auto." (English: "The man drives a yellow car.") with a narrow focus on either on the subject ("Mann"), the adjective ("gelbes"), or the object ("Auto") was used. Such simple sentences of the structure subject, verb, adjective, object were previously applied in a prosody test-battery (Meister et al., 2009).

As an example, for assessment of a narrow focus on the adjective the dialogue was as follows:

Interlocutor 1 (stating): "The man drives a yellow car."

Interlocutor 2 (mirroring): "The man drives a yellow car."

Interlocutor 1 (stating): "The man drives a *red* car."

Interlocutor 2 (correcting): "The man drives a YELLOW car."

This last sentence of interlocutor 2 was taken as stimulus. A similar procedure was applied for the other two focus conditions ("MANN", "AUTO"). Such a dialogue situation was assumed to give more realistic facial movements and head gestures than, for instance, reading the sentence aloud (Cvejic et al., 2010). Two lab members acted as the interlocutors. During this procedure the two interlocutors were seated face-to-face at a distance of about 1.4 m. About 60 cm in front of interlocutor 2 an apple[TM] iPhone 13 mini was placed on a small tripod. It was used for video-recording the face of interlocutor 2 and also provided the audio signal.

The video was then played back on a monitor (hp e243i, 24''). Head and facial movements were registered via the "live face" app (Reallusion[TM] Inc., Atlanta, USA) running on the iPhone, which was placed 40 cm in front of the monitor. Figure 1 gives an example of the facial motion tracking. Alternatively, the mocap data could also have been collected "live" during the dialogue. However, recording the video first and then applying motion capture had the practical advantage that the video and the corresponding animation could be compared directly, which is essential for the technical evaluation described here.

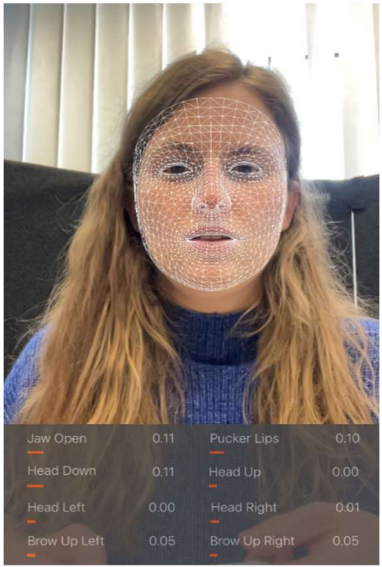

Figure 1: Head and facial motion capture provided by the "live face" app (Reallusion<sup>TM</sup>) running on an iPhone 13 mini.

*Animating head and face movements in a virtual human*

The mocap data recorded by the "live face" app were streamed to a PC (Intel Core i7-3770, 3.4 GHz, 4 kernel, 16 GB RAM, Win10) via USB tethering. On this PC, the data were processed by means of the "Motion LIVE" plugin which was installed within the iClone7 software package (Reallusion<sup>TM</sup> Inc., Atlanta, USA). iClone7 is a real-time animation software, which allows a video-realistic scene design using virtual humans. This software environment offers the opportunity of parametrizing and arbitrarily manipulating head and facial elements such as eyebrows, eyes and mouth. The manipulations can be achieved either separately for each of the different elements or globally by adjusting the "expression strength" with values between 0% (no head and facial movements) and 200% (maximum head and facial movements). In addition, the iClone7 software offers visualizing articulatory movements by using the "Accu-lips" plugin and thus additionally enables generating and modifying visual speech cues in a virtual human.

We used the "Motion LIVE" plugin to animate head and facial movements of a virtual human on the basis of the mocap data provided by the "live face" app. Additionally, using the "Accu-lips" plugin, visemes were manually aligned with the utterance, which allowed to lip-synchronize the audio signal and the articulatory movements. This procedure resulted in a video-realistic representation of the phrases recorded during the dialogue situation described above. Figure 2 shows the virtual human at different time points when articulating a sentence with a parameter setting of 150%, as an example.

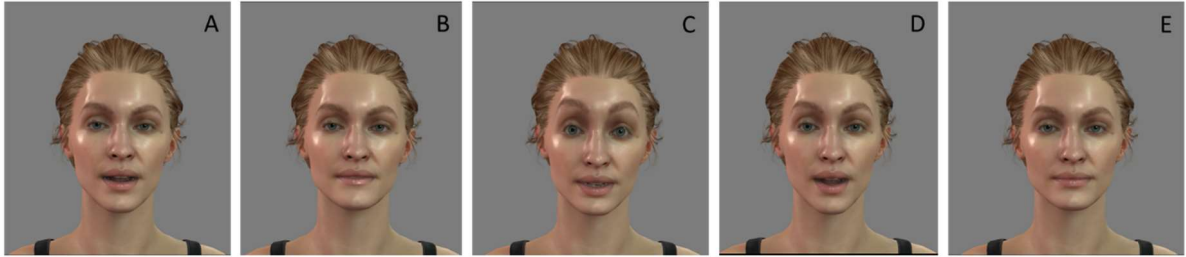

Figure 2: Virtual human used for animation of head and face movements. The stills A-E show the virtual human at different time points in chronological order, when uttering the sentence "Der Mann faehrt ein GELBES Auto." The phase of the narrow focus ("GELBES") is covered by C. The "expression strength" was set to 150%.

*Evaluation of head and eyebrow movements*

In order to compare the head and eyebrow movements between the video of the real talker and the virtual human, techniques of automated facial landmark localization were used. For each video frame, the dlib library (King, 2009) was used to recognize the face and to detect 68 face landmarks that represent characteristic 2D points, including eyebrows, eyes, nose, mouth, and jaw line. Landmark coordinates were temporally smoothed with a Savitzky-Golay filter (window length = 13 frames, polynomial order = 2) in order to reduce jitter between successive frames. We analyzed the head movement – in particular the *pitch angle* of the vertical head rotation (nodding) – on the basis of a subset of six face landmarks by means of the Perspective-n-Point solving algorithm provided by OpenCV (Bradski, 2000). In addition, we analyzed the *eyebrow raise* by determining the Euclidean distance between the landmarks at the medial palpebral commissure ("inner corner") and the innermost point of the eyebrow of the right eye, as tracking was found to be most robust for these landmarks. The influence of perspective distortion on the eyebrow movement estimation due to varying vertical head rotation was compensated by dividing the measured eyebrow raise by the cosine of the pitch angle. In order to give a common metric, the outcome of the movement estimation (in pixels) was finally calibrated to millimeters by measuring the distance between the eyes of the real talker and assuming the same size for the virtual human.

**Results and discussion**

Figure 3 displays the example "Der Mann faehrt ein gelbes Auto." with a narrow focus either on the subject ("Mann"; A), the adjective ("gelbes"; B), or the object ("Auto"; C) of the sentence. The top panels show the waveform, the F0-contour (blue line) and the intensity (green line) of the speech signal. In each case, the stressed word is clearly detectable based on the display of the acoustic cues, demonstrated by an increase in both F0 and the intensity.

The middle panels show the corresponding head rotation in the vertical plane ("pitch"). When looking at the movements of the real talker (dashed line), the critical phase of the narrow focus is typically preceded by raising the head and accompanied by a pronounced downward pitch. This pattern is generally similar to what has been described previously, though movements reveal talker idiosyncrasies (see Kim et al., 2014, and Wagner et al., 2014 for an overview).

Regarding the head rotations of the virtual human, it appears that they closely follow the time course of the real talker. Calculating Pearson's correlation coefficients between the real and the simulated movements gives values between 0.80 and 0.98 (see Table 1). In terms of the magnitude, the downward movements are best approximated with adjusting the "expression strength" to 200%. Based on this parameter, the panels show that manipulating the expression strength has a linear impact on the magnitude of head rotations. An estimation reveals that each 50% change in "expression strength" results in a pitch change of about 1.5 -2 degrees. A choice of 0% (not shown) completely eliminates all rigid head movements.

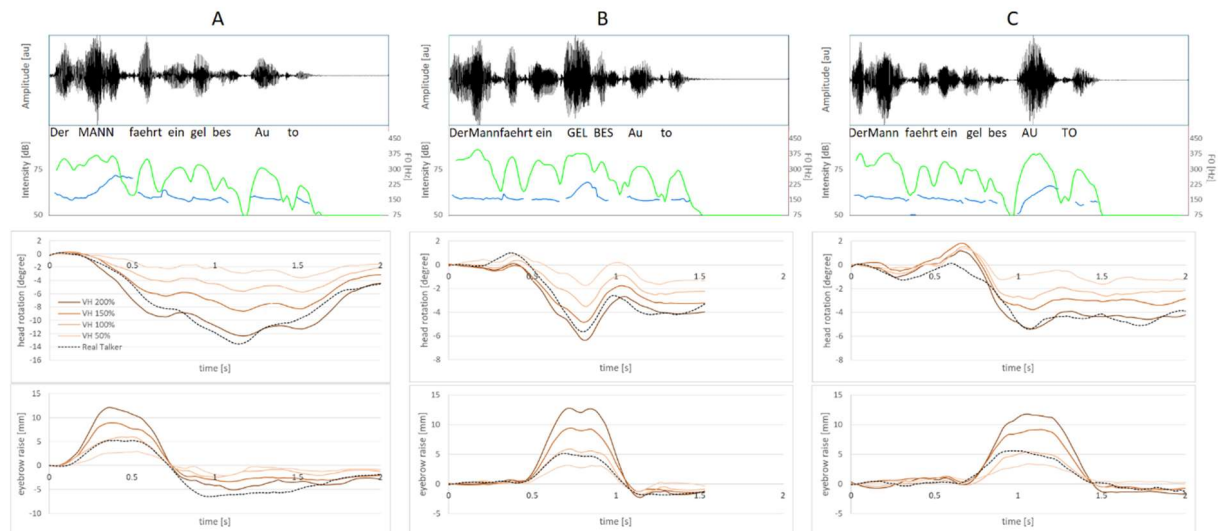

Figure 3: Phrase in the three narrow focus conditions (stress on "MANN" (A), "GELBES" (B), or "AUTO"(C)). Top panels: waveform, F0 contour (blue) and intensity contour (green) of the utterances. Middle panels: head rotation (pitch angle). Bottom panels: eyebrow movements. The dashed line shows the movements of the real talker, the solid lines show the movements of the virtual human (VH). Different "expression strengths", varying between 50% and 200% are given. Head rotation and eyebrow movement were normalized to the first frame (t=0 s).

The bottom panels show the eyebrow movements of the real talker (dashed line) and the virtual human. The results of both the real talker and the virtual human are generally in line with previous studies, reporting eyebrow raises prior and during the critical phase and a subsequent release (Kim et al., 2014, Wagner et al., 2014). Similarly, the movements of our real talker and the virtual human are comparable, resulting in correlation coefficients between 0.92 and 0.98 (see Table 1). Eyebrow movements of the virtual talker with 100% expression strength closely match those of the real talker. As with the head rotation, the effect of parametrization is clearly visible. The effect can be estimated by a change of about 2-3 mm per 50% change in expression strength. Again, a choice of 0% (not shown) completely eliminates all eyebrow movements.

|  | Head rotation | | | | | Eyebrow raise | | | |
|---|---|---|---|---|---|---|---|---|---|
| Strength [%] | 200 | 150 | 100 | 50 | | 200 | 150 | 100 | 50 |
| Focus "MANN" | 0.97 | 0.97 | 0.95 | 0.88 | | 0.95 | 0.95 | 0.94 | 0.92 |
| Focus "GELBES" | 0.97 | 0.96 | 0.95 | 0.87 | | 0.97 | 0.97 | 0.98 | 0.94 |
| Focus "AUTO" | 0.98 | 0.96 | 0.93 | 0.80 | | 0.96 | 0.97 | 0.96 | 0.94 |

Table 1: Pearson's correlation coefficients showing a strong relationship between the movements of the real talker and the virtual human for "expression strengths" between 50 and 200%. All p<0.001.

Taken together, the technical evaluation of the proposed method shows that rigid head and eyebrow movements of the virtual human are qualitatively similar to those of the video-recorded real talker. The critical period of the narrow focus is preceded by eyebrow and head raises, and it is accompanied by a pronounced downward head rotation. In terms of the quantitative evaluation it appears that eyebrow movements of the virtual human are closely related to those of the real talker, as confirmed by the high correlation coefficients. The fact that the correlations typically decrease with lower expression strength (see Table 1) is due to the strongly reduced movements, especially with a parameter choice of 50%. However, compared to the eyebrow raise, the magnitude of head movements is more restricted in the virtual human, as a strength of 200% had to be chosen to match the real talker. Still, it should be noted that only the global parametrization of the virtual human was considered and further adjustment is possible via manual tuning.

The option of parametrizing single elements such as head rotation, eyebrow raises, and articulatory movements constitutes a valuable research tool. Our evaluation of the global expression strength adjustment has shown that a targeted modification of the movements is possible. Hence, the proposed method is generally suitable for investigating multimodal aspects of communication behavior in a realistic and ecologically valid manner and opens up new avenues for examining aspects of verbal and nonverbal interaction. Our current work uses this procedure for the examination of audiovisual prosody perception in CI users. These individuals have a restricted spectro-temporal resolution because of the limited number of electrodes and the current spread associated with the propagation of the electric field (cf., Friesen et al., 2001). A number of studies have shown that especially recognizing changes in the intonation pattern based on F0 is limited (cf. Meister et al., 2011). However, since intonation is coupled with facial expressions and head movements, *multimodal* prosody perception in CI users might be less restricted, especially because listeners with profound hearing loss often reveal enhanced speechreading abilities (cf., Strelnikov et al., 2009; Layer et al., 2022).

Unfortunately, only a few studies have tapped into this topic so far. Recently, Lasfargues-Delannoy et al. (2021) reported supra-normal skills in processing of visuo-auditory prosodic information in CI listeners. These authors compared the discrimination of questions and statements in the auditory, visual and audiovisual modality between CI users and normal-hearing listeners presented with vocoder simulations of cochlear implants. Videos of two actors were generated as stimuli. In order to avoid ceiling effects in the visual modality, the actors were instructed to reduce their facial expressions. Compared to the discrimination performance in the best single modality (i.e., auditory or visual) the CI recipients but not the NH listeners showed a significant benefit when both modalities were combined. This was despite the fact that – contrary to the expectation – the CI users performed significantly worse than the NH listeners in the visual-only condition. The authors found that the CI users tended to look more at the mouth of the actors whereas age-matched normal-hearing listeners also explored the eye-region. However, since the study used video-recordings of actors, it was not possible to disentangle effects based on different face regions.

This is where the proposed method based on virtual humans is able to give additional information and promises to uncover multimodal effects of prosody perception. As shown above, the option of providing video-realistic prosody cues and parametrizing them arbitrarily avoids ceiling effects on the one hand and also allows animating isolated mouth and facial cues based on a common underlying stimulus on the other hand. Such examinations are currently underway, possibly giving novel information on multimodal facilitation in CI listeners relevant to everyday communication.

**Acknowledgements**

Supported by the Deutsche Forschungsgemeinschaft (ME 2751-4.1) to HM and (SA 3615/1-1 and SA 3615/2-1) to PS.